\documentclass{article}

\usepackage{arxiv}

\usepackage[utf8]{inputenc} 
\usepackage[T1]{fontenc}    
\usepackage{hyperref}       
\usepackage{url}            
\usepackage{booktabs}       
\usepackage{amsfonts}       
\usepackage{nicefrac}       
\usepackage{microtype}      
\usepackage{lipsum}
\usepackage{multicol}
\usepackage{graphicx}
\usepackage{xcolor}
\graphicspath{{media/}}
\title{Transformers predicting the future. Applying attention in next-frame and time series forecasting.}

\author{
 \textbf{Radostin Cholakov} \\
  High School of Mathematics\\
  "Acad. Kiril Popov" - Plovdiv, Bulgaria \\
  \texttt{rsg.group.here@gmail.com} \\
\and
 \textbf{Todor Kolev} \\%
  Comrade Cooperative\\%
  Sofia, Bulgaria\\%
  \texttt{t.kolev@comrade.coop}%
}

\begin{document}
\maketitle
\begin{abstract}
Recurrent Neural Networks were, until recently, one of the best ways to capture the timely dependencies in sequences. However, with the introduction of the Transformer, it has been proven that an architecture with only attention-mechanisms without any RNN can improve on the results in various sequence processing tasks (e.g. NLP). Multiple studies since then have shown that similar approaches can be applied for images, point clouds, video, audio or time series forecasting. Furthermore,  solutions such as the Perceiver or the Informer have been introduced to expand on the applicability of the Transformer. Our main objective is testing and evaluating the effectiveness of applying Transformer-like models on time series data, tackling susceptibility to anomalies, context awareness and space complexity by fine-tuning the hyperparameters, preprocessing the data, applying dimensionality reduction or convolutional encodings, etc. We are also looking at the problem of next-frame prediction and exploring ways to modify existing solutions in order to achieve higher performance and learn generalized knowledge.
\end{abstract}

\vskip 0.6cm


\begin{multicols}{2}

\section{Introduction}

Since its introduction, the Transformer \cite{vaswani2017attention} has revolutionized how neural networks can process sequential data and is currently the go to solution for a wide variety of natural language processing tasks. Analogous models are also being applied for image and video, point clouds \cite{jaegle2021perceiver}, sound and time series data.

The Transformer adopts an encoder-decoder structure where the core function of each encoder layer is to generate information about which parts of the inputs are relevant to each other. The decoder part does the opposite, taking all the encodings and using their incorporated contextual information to generate an output sequence.

The inputs and outputs (target sequences) are first embedded into an n-dimensional space and since there are no recurrent networks that can remember how sequences are fed into a model the positions are added to the embedded representation of each item.

Both the encoder and the decoder are composed of modules that can be stacked on top of each other multiple times. The modules consist mainly of Multi-Head Attention and Feed Forward layers.

In this context, the attention-mechanism can be described as mapping a query and a set of key-value pairs to an output.

\begin{equation}
    Attention(Q, K, V) = softmax(\frac{QK^T}{\sqrt{d_k}})V
\end{equation}

$Q$ is a matrix containing the query (vector representation of one item in the sequence), $K$ are all the keys (vector representations of all the items in the sequence) of dimension $d_k$ and $V$ are the values of dimension $d_v$. This means that the weights are defined by how each item of the sequence ($Q$) is influenced by all the other items in the sequence ($K$). Additionally, the SoftMax\footnote{\url{https://en.wikipedia.org/wiki/Softmax_function}} function is applied produce a distribution between 0 and 1. Those weights are then applied to all the words in the sequence that are introduced in $V$. They are the same vectors as $Q$ for the first attention blocks in the encoder and decoder (self attention) but different for the module that has both encoder and decoder inputs (cross attention).

Instead of performing a single attention function the multi-head attention linearly projects the queries, keys and values $h$ times in parallel.

After the multi-attention heads in both the encoder and decoder, there are pointwise feed-forward layers having identical parameters for each position, which can be described as a separate, identical linear transformation of each element from the given sequence.

In recent studies it has been shown that similar approaches could lead to significant performance boosts in tasks other than NLP. The Vision Transformer (ViT), \cite{dosovitskiy2020image}, attains excellent results in computer vision compared to state-of-the-art convolutional networks while requiring substantially fewer computational resources to train. Solutions such as the VideoGPT, \cite{yan2021videogpt}, showcase how to efficiently apply Transformers for video generation tasks. In the field of time series forecasting there are multiple proposals \cite{li2019enhancing} on how Transformers can be modified to compensate for their susceptibility to anomalies while simultaneously leveraging the performance advantages.

With all this as a context we will examine if and how Transformers can be used for predicting future events, going from traditional approaches with time series data (e.g. weather or stock price forecasting) to more abstract tasks such as next-frame prediction in a video where the model should learn different movement patterns and additional dependencies.

\section{Transformers for time series forecasting}

Time series forecasting plays an important role in daily life to help people manage resources and make decisions. Although still widely used, taditional models, such as State Space Models \cite{durbin2012time} and Autoregressive\footnote{\url{https://en.wikipedia.org/wiki/Autoregressive_model}} (AR) models, are designed to fit each time series independently and require practitioners’ expertise in manually selecting trend, seasonality, etc. To tackle those challenges, recurrent neural networks \cite{salinas2020deepar} have been proposed as an alternative solution. Despite the emergence of various variants, including LSTM \cite{hochreiter1997long} and GRU \cite{shen2018deep}, it is still hard to capture long-term dependencies in TS data. Unlike the RNN-based methods, Transformers allow the model to access any part of the history regardless of distance, making it potentially more suitable for grasping the recurring patterns with long-term dependencies.

\subsection{Challanges and Solutions}

As described in \cite{NEURIPS2019_6775a063}, Transformers give impressive results for their performance advantages in forecasting tasks. However, their self-attention matches queries against keys insensitive to local context, which may make the model prone to anomalies and bring underlying optimization issues. Whether an observed point is an anomaly, change point or part of the patterns is dependent on its surrounding context. The similarities between queries and keys are computed based on their point-wise values without fully taking into account local context. In previous studies convolutional self-attention\footnote{\url{https://github.com/mlpotter/Transformer_Time_Series}} has been proposed to ease the issue.

Another issue which may emerge is related to the space complexity of canonical Transformer which grows quadratically with the input length $L$, causing memory bottleneck. 

Solutions such as the \textit{Sparse Transformer} \cite{child2019generating}, with complexity of $O(n\sqrt{n})$ and the \textit{LogSparse Transformer} \cite{NEURIPS2019_6775a063}, with complexity reduced to $O(n(\log{n})^2)$ have been introduced. These approaches make long time series modeling feasible while retaining comparable to canonical Transformer results with much less memory usage.

\subsection{Experiments and Results}

\begin{figure}[h!]
  \includegraphics[width=\textwidth]{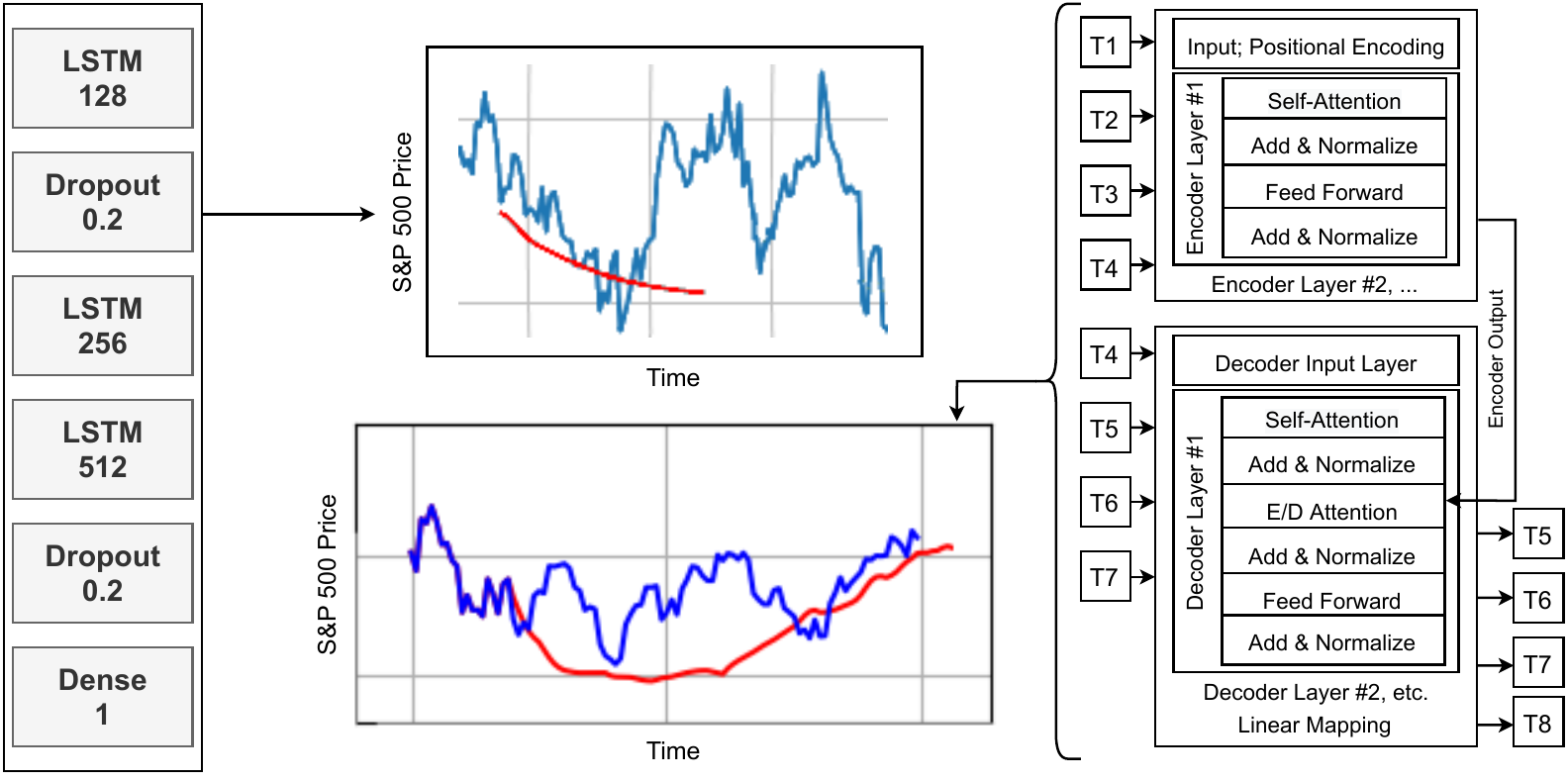}
  \caption{S\&P500 price forecasting experiment. The graphics on the left and on the right describe the architectures. The top price chart shows how the RNN LSTM model forecasts pricing. The bottom chart shows the Transformer's predictions. Forecasts are colored in red and the actual prices - in blue.}
  \label{fig:sp500}
\end{figure}

During our research we compared two architectures - a standard RNN utilizing LSTM cells and a simple implementation of a Transformer (See Fig. \ref{fig:sp500}) for forecasting how the price of the S\&P500 index \footnote{\url{https://en.wikipedia.org/wiki/S\%26P_500}} will change. They were trained on the same amount of data\footnote{The data, some of the code, models and experiments described in this study are available on \url{https://github.com/radi-cho/SRS21-public-data}}: the daily closing value of the index from \textit{3rd Jan 2000} to \textit{31st Aug 2018}.

As shown in Fig. \ref{fig:sp500} the LSTM recurrent neural network barely learns to follow a trend whereas the Transformer architecture is able to capture more detailed dependencies and use them for future forecasting. \textit{For example: In the short-term, the index price usually goes up after the quarterly reports of the big companies in a good year.}

\end{multicols}

\begin{multicols}{2}

\section{Transformers for next-frame prediction}

Another, more abstract way of thinking about future forecasting is next-frame prediction \cite{zhou2020deep}. That is, predicting what happens next in the form of newly generated images, after a given amount of historical images. It refers to starting from continuous, unlabeled video frames and constructing a network that can accurately generate subsequent frames. Next-frame prediction is not only an experimental approach for video processing but a gateway to modelling machine learning architectures that can do more general assumptions and abstract reasoning.

\subsection{Methods}

The introduction of GPT and Image-GPT, \cite{chen2021pre} - a class of autoregressive Transformers that have shown tremendous success in modelling discrete data, inspired the creation of more and more Transformer-like solutions specialized for different tasks. As a part of the research we examined the VideoGPT, \cite{yan2021videogpt}, a conceptually simple architecture for scaling likelihood based generative modeling to videos.

\begin{figure}[h!]
  \begin{center}
      \includegraphics[width=10cm]{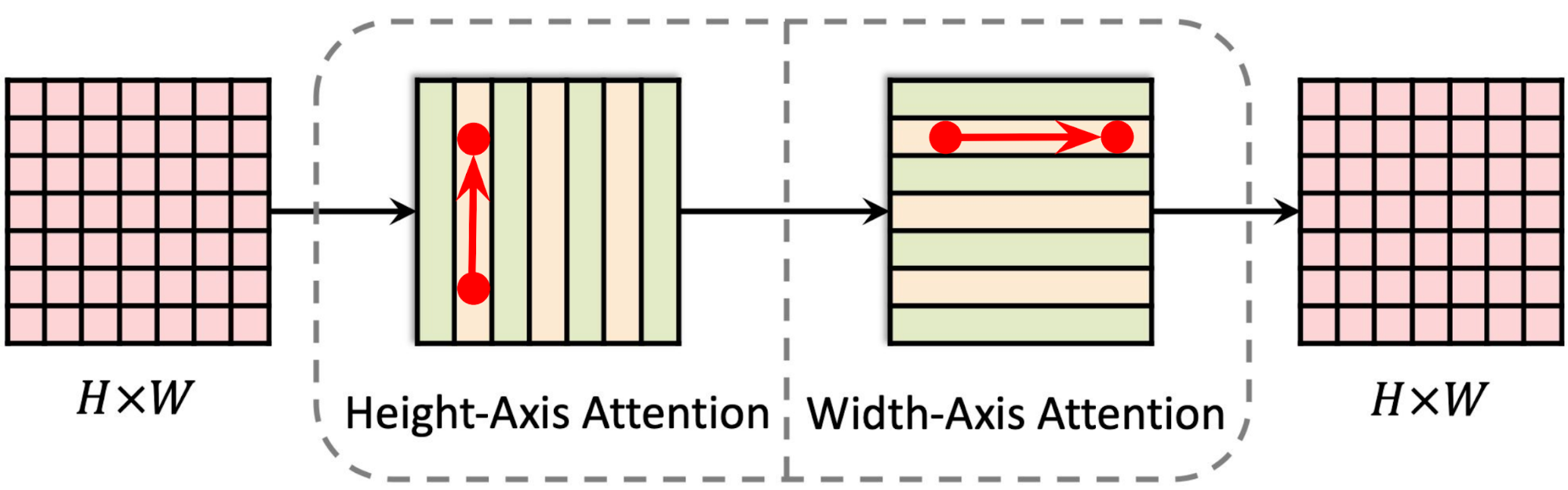}
  \end{center}
  \caption{Axial Attention. The vertical layer provides 1-dimensional self-attention globally, propagating information within individual columns while the horizontal 1D layer allows for the capture of column-wise as well as row-wise information. That way the complexity of self-attention is reduced from quadratic (2D) to linear (1D).}
  \label{fig:axial}
\end{figure}

VideoGPT uses Vector Quantized Variational Autoencoder (VQ-VAE) \cite{razavi2019generating} to learn downsampled latent representations of a given video. It employs 3D convolutions and axial self-attention \cite{ho2019axial} - generalization of self-attention that naturally aligns with the multiple dimensions of the tensors in both the encoding and the decoding settings. It allows for the vast majority of the context to be computed in parallel during decoding (Fig. \ref{fig:axial}).

\begin{figure}[h!]
  \includegraphics[width=\textwidth]{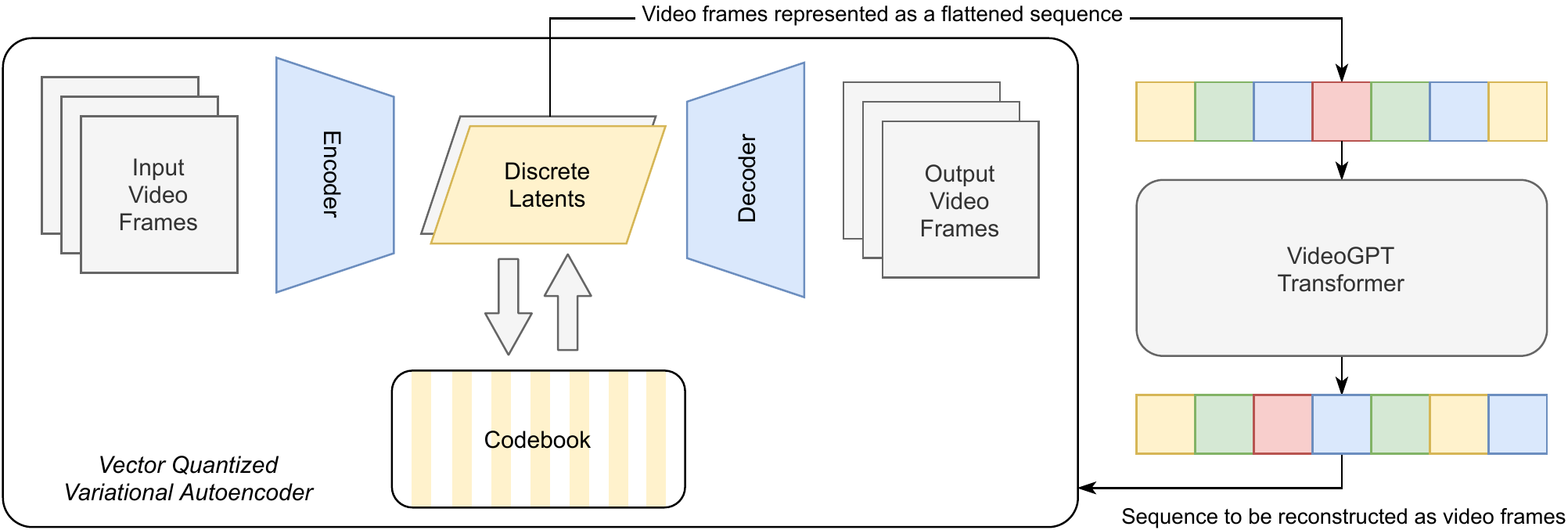}
  \caption{VideoGPT. The training pipeline is broken into two sequential stages. The first stage (Left) is similar to the original VQ-VAE training procedure. During the second stage (Right), VQ-VAE encodes video data to latent sequences as training data for the prior model.}
  \label{fig:videogpt}
\end{figure}

A simple GPT-like architecture  (Fig. \ref{fig:videogpt}) is then used to autoregressively model the discrete latents using position encodings.

\end{multicols}

\begin{multicols}{2}

Despite the simplicity and ease of training, the VideoGPT is able to generate samples competitive with state-of-the-art GAN \cite{goodfellow2014generative} models. The experiments in the original paper mainly focus on creative video generation where the model samples a single frame and then tries to guess what the video is about. Although the VQ-VAE is trained fully unconditionally, conditional samples are still possible by training a conditional prior.

The model was modified and retrained to condition $N$ frames and produce $N + M$ frames while decoding with a VQ-VAE trained with sequence length $N + M$. The conditioned frames are firstly fed into a 3D ResNet\footnote{\url{https://towardsdatascience.com/an-overview-of-resnet-and-its-variants-5281e2f56035}} \cite{he2016deep}, and then cross-attention is performed on the ResNet output during prior network training.

For the experiments we used a composition of two moving\footnote{Pre-generated open-source moving MNIST datasets, such as \url{http://www.cs.toronto.edu/~nitish/unsupervised_video/} introduced in \cite{srivastava2015unsupervised} already exist. For our study moving videos were generated from the static MNIST images with a Python script in order to capture additional labelling information, e.g. which numbers are shown in the video, if they collide with one another, etc. For reference: \url{https://gist.github.com/tencia/afb129122a64bde3bd0c}.} MNIST, \cite{lecun-mnisthandwrittendigit-2010}, handwritten digits in a square box with dimensions of 64x64 pixels. They can bounce off the walls and go over each other.

Our main objective is to run the conditioned model and generate a few seconds (usually 4, 8, 16 or 32 frames) of video predicting the change of the position of the handwritten digits. Furthermore, if successful, the architecture can be repurposed and instead of feeding the result encodings to the VQ-VAE decoder, they can be passed to an additional classification neural network which performs labelling on the predicted future of the video. For example: to classify how likely it is the digits to collide in the next $M$ frames.

\subsection{Results}

\begin{figure}[h!]
  \begin{center}
      \includegraphics[width=12cm]{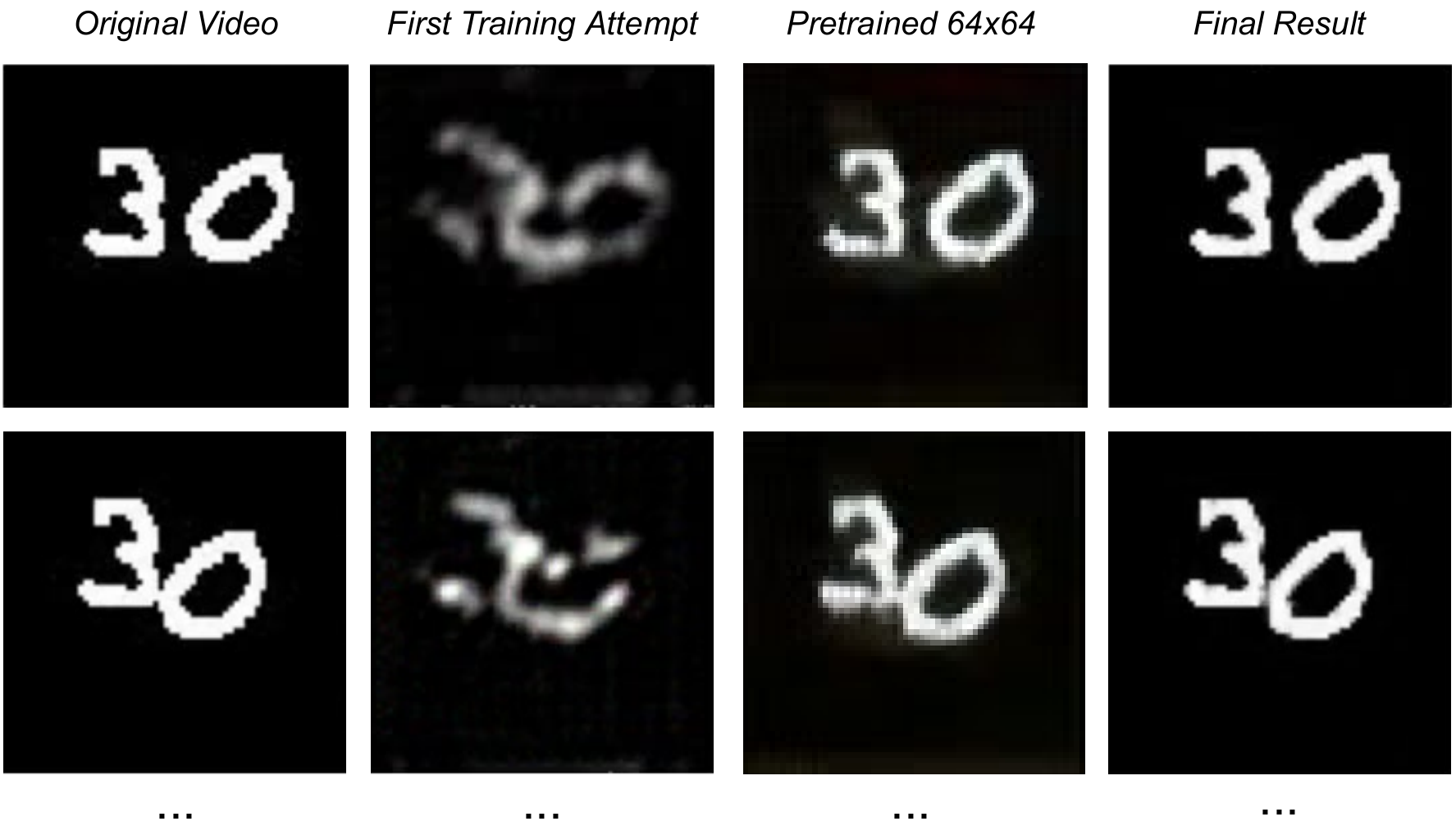}
  \end{center}
  \caption{Representations of the original video from VQ-VAE.}
  \label{fig:mnist-vqvae}
\end{figure}

We have successfully trained a VQ-VAE with sequence lengths of 4, 8, 16 and 32 frames on the generated moving MNIST database. The final decoder reconstructions are more accurate than the pretrained models mentioned in the original paper (Fig \ref{fig:mnist-vqvae}).

\end{multicols}

\begin{figure}[h!]
  \includegraphics[width=\textwidth]{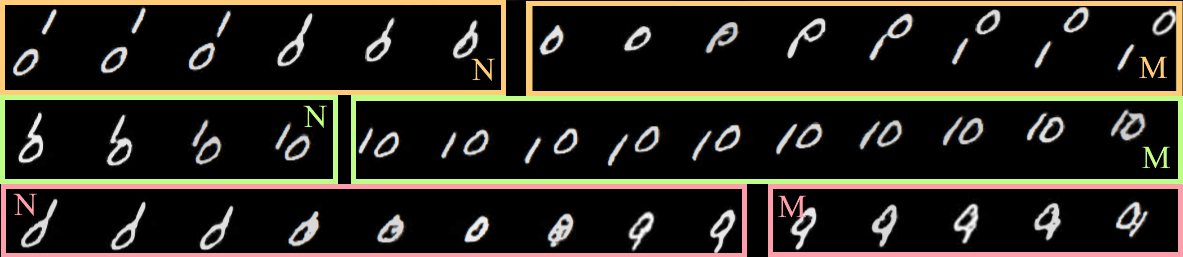}
  \caption{Next-frame prediction in a Moving MNIST video. Conditioning the first $N$ frames and generating a video of length $N + M$ frames, where the $M$ frames are newly generated.}
  \label{fig:mnist-condition}
\end{figure}

\begin{multicols}{2}

Moreover, the modified VideoGPT instance which predicts multiple frames in the future by conditioning historical data has been successful in the task of forecasting moving MNIST videos (Fig. \ref{fig:mnist-condition}). It has been tested for sequences of 4 (condition 2 frames to predict the next 2), 8 (condition 4 to predict 4) and 16 (condition 8 to predict 8).

Both the results from the VideoGPT and the time series experiments are a proof that with certain modifications the Transfromer can lead to high accuracy predictions and can replace traditional methods such as RNNs and CNNs in the field of future forecasting.

\section{Future Work}

We have shown how the generative VideoGPT \cite{yan2021videogpt} model can be tailored for future frame predictions as well as additional classification tasks. One of the directions for future development is to polish up those proposals and clear out some of the assumptions made in this paper to end up with a more stable and predictable architecture. In various studies (e.g. \cite{wu2021cvt, dehghani2018universal, liu2020convtransformer}), convolutions or recurrent structures have been used in order to manipulate, preprocess the encodings and enhance the Transformer's performance. We're interested in similar approaches in combination with the encoder/decoder weights of a Transformer to be able to process concepts with higher levels of complexity.

If the two digits in the VideoGPT experiments will collide depends on wether they're going towards one another, but if the model is able to predict that they will bounce off a wall and then change directions to eventually collide is another level of abstraction and reasoning. Additionally, instead of videos the proposed pipeline can be used for encoded representations of other data types - time series, etc.

Another fruitful direction would be to explore the effectiveness of the Transformer as a part of Deep Reinforcement Learning\footnote{\url{https://en.wikipedia.org/wiki/Deep_reinforcement_learning}} environments. We will be looking at already existing research (e.g. \cite{zambaldi2018deep, chen2021decision, team2021open}) in order to come up with efficient solutions combining advantages from the worlds of supervised and reinforcement learning.

\section{Conclusion}

We have presented multiple ways of forecasting the future and how Transformer-like architectures can be adopted for such an use. We have looked at the possible solutions to problems emerging when Transformers are applied to time series data and the different levels of abstraction they can perform. RNNs and other standard solutions have been compared to newly introduced models. We have also modified the VideoGPT model to be used conditionally for next-frame prediction and proposed ways to upgrade it for future classification tasks and general reasoning. It can even be integrated as a part of Reinforcement Learning environments to enhance the behaviour of RL agents. We hope that our work done during the Summer Research School 2021 in Apriltsi, Bulgaria will be useful for future design of architectures in time series forecasting, video generation, decision making, etc.

\nocite{*}
{\small \bibliography{references} }

\begin{thebibliography}{10}

\bibitem{vaswani2017attention}
Ashish Vaswani, Noam Shazeer, Niki Parmar, Jakob Uszkoreit, Llion Jones,
  Aidan~N Gomez, {\L}ukasz Kaiser, and Illia Polosukhin.
\newblock Attention is all you need.
\newblock In {\em Advances in neural information processing systems}, pages
  5998--6008, 2017.

\bibitem{jaegle2021perceiver}
Andrew Jaegle, Felix Gimeno, Andrew Brock, Andrew Zisserman, Oriol Vinyals, and
  Joao Carreira.
\newblock Perceiver: General perception with iterative attention.
\newblock {\em arXiv preprint arXiv:2103.03206}, 2021.

\bibitem{dosovitskiy2020image}
Alexey Dosovitskiy, Lucas Beyer, Alexander Kolesnikov, Dirk Weissenborn,
  Xiaohua Zhai, Thomas Unterthiner, Mostafa Dehghani, Matthias Minderer, Georg
  Heigold, Sylvain Gelly, et~al.
\newblock An image is worth 16x16 words: Transformers for image recognition at
  scale.
\newblock {\em arXiv preprint arXiv:2010.11929}, 2020.

\bibitem{yan2021videogpt}
Wilson Yan, Yunzhi Zhang, Pieter Abbeel, and Aravind Srinivas.
\newblock Videogpt: Video generation using vq-vae and transformers.
\newblock {\em arXiv preprint arXiv:2104.10157}, 2021.

\bibitem{li2019enhancing}
Shiyang Li, Xiaoyong Jin, Yao Xuan, Xiyou Zhou, Wenhu Chen, Yu-Xiang Wang, and
  Xifeng Yan.
\newblock Enhancing the locality and breaking the memory bottleneck of
  transformer on time series forecasting.
\newblock {\em Advances in Neural Information Processing Systems},
  32:5243--5253, 2019.

\bibitem{durbin2012time}
James Durbin and Siem~Jan Koopman.
\newblock Time series analysis by state space methods oxford university press,
  2012.

\bibitem{salinas2020deepar}
David Salinas, Valentin Flunkert, Jan Gasthaus, and Tim Januschowski.
\newblock Deepar: Probabilistic forecasting with autoregressive recurrent
  networks.
\newblock {\em International Journal of Forecasting}, 36(3):1181--1191, 2020.

\bibitem{hochreiter1997long}
Sepp Hochreiter and J{\"u}rgen Schmidhuber.
\newblock Long short-term memory.
\newblock {\em Neural computation}, 9(8):1735--1780, 1997.

\bibitem{shen2018deep}
Guizhu Shen, Qingping Tan, Haoyu Zhang, Ping Zeng, and Jianjun Xu.
\newblock Deep learning with gated recurrent unit networks for financial
  sequence predictions.
\newblock {\em Procedia computer science}, 131:895--903, 2018.

\bibitem{NEURIPS2019_6775a063}
Shiyang Li, Xiaoyong Jin, Yao Xuan, Xiyou Zhou, Wenhu Chen, Yu-Xiang Wang, and
  Xifeng Yan.
\newblock Enhancing the locality and breaking the memory bottleneck of
  transformer on time series forecasting.
\newblock In H.~Wallach, H.~Larochelle, A.~Beygelzimer, F.~d\textquotesingle
  Alch\'{e}-Buc, E.~Fox, and R.~Garnett, editors, {\em Advances in Neural
  Information Processing Systems}, volume~32. Curran Associates, Inc., 2019.

\bibitem{child2019generating}
Rewon Child, Scott Gray, Alec Radford, and Ilya Sutskever.
\newblock Generating long sequences with sparse transformers.
\newblock {\em arXiv preprint arXiv:1904.10509}, 2019.

\bibitem{zhou2020deep}
Yufan Zhou, Haiwei Dong, and Abdulmotaleb El~Saddik.
\newblock Deep learning in next-frame prediction: a benchmark review.
\newblock {\em IEEE Access}, 8:69273--69283, 2020.

\bibitem{chen2021pre}
Hanting Chen, Yunhe Wang, Tianyu Guo, Chang Xu, Yiping Deng, Zhenhua Liu, Siwei
  Ma, Chunjing Xu, Chao Xu, and Wen Gao.
\newblock Pre-trained image processing transformer.
\newblock In {\em Proceedings of the IEEE/CVF Conference on Computer Vision and
  Pattern Recognition}, pages 12299--12310, 2021.

\bibitem{razavi2019generating}
Ali Razavi, Aaron van~den Oord, and Oriol Vinyals.
\newblock Generating diverse high-fidelity images with vq-vae-2.
\newblock In {\em Advances in neural information processing systems}, pages
  14866--14876, 2019.

\bibitem{ho2019axial}
Jonathan Ho, Nal Kalchbrenner, Dirk Weissenborn, and Tim Salimans.
\newblock Axial attention in multidimensional transformers.
\newblock {\em arXiv preprint arXiv:1912.12180}, 2019.

\bibitem{goodfellow2014generative}
Ian Goodfellow, Jean Pouget-Abadie, Mehdi Mirza, Bing Xu, David Warde-Farley,
  Sherjil Ozair, Aaron Courville, and Yoshua Bengio.
\newblock Generative adversarial nets.
\newblock {\em Advances in neural information processing systems}, 27, 2014.

\bibitem{he2016deep}
Kaiming He, Xiangyu Zhang, Shaoqing Ren, and Jian Sun.
\newblock Deep residual learning for image recognition.
\newblock In {\em Proceedings of the IEEE conference on computer vision and
  pattern recognition}, pages 770--778, 2016.

\bibitem{srivastava2015unsupervised}
Nitish Srivastava, Elman Mansimov, and Ruslan Salakhudinov.
\newblock Unsupervised learning of video representations using lstms.
\newblock In {\em International conference on machine learning}, pages
  843--852. PMLR, 2015.

\bibitem{lecun-mnisthandwrittendigit-2010}
Yann LeCun and Corinna Cortes.
\newblock {MNIST} handwritten digit database.
\newblock 2010.

\bibitem{wu2021cvt}
Haiping Wu, Bin Xiao, Noel Codella, Mengchen Liu, Xiyang Dai, Lu~Yuan, and Lei
  Zhang.
\newblock Cvt: Introducing convolutions to vision transformers.
\newblock {\em arXiv preprint arXiv:2103.15808}, 2021.

\bibitem{dehghani2018universal}
Mostafa Dehghani, Stephan Gouws, Oriol Vinyals, Jakob Uszkoreit, and {\L}ukasz
  Kaiser.
\newblock Universal transformers.
\newblock {\em arXiv preprint arXiv:1807.03819}, 2018.

\bibitem{liu2020convtransformer}
Zhouyong Liu, Shun Luo, Wubin Li, Jingben Lu, Yufan Wu, Chunguo Li, and Luxi
  Yang.
\newblock Convtransformer: A convolutional transformer network for video frame
  synthesis.
\newblock {\em arXiv preprint arXiv:2011.10185}, 2020.

\bibitem{zambaldi2018deep}
Vinicius Zambaldi, David Raposo, Adam Santoro, Victor Bapst, Yujia Li, Igor
  Babuschkin, Karl Tuyls, David Reichert, Timothy Lillicrap, Edward Lockhart,
  Murray Shanahan, Victoria Langston, Razvan Pascanu, Matthew Botvinick, Oriol
  Vinyals, and Peter Battaglia.
\newblock Deep reinforcement learning with relational inductive biases.
\newblock In {\em International Conference on Learning Representations}, 2019.

\bibitem{chen2021decision}
Lili Chen, Kevin Lu, Aravind Rajeswaran, Kimin Lee, Aditya Grover, Michael
  Laskin, Pieter Abbeel, Aravind Srinivas, and Igor Mordatch.
\newblock Decision transformer: Reinforcement learning via sequence modeling.
\newblock {\em arXiv preprint arXiv:2106.01345}, 2021.

\bibitem{team2021open}
Ended~Learning Team, Adam Stooke, Anuj Mahajan, Catarina Barros, Charlie Deck,
  Jakob Bauer, Jakub Sygnowski, Maja Trebacz, Max Jaderberg, Michael Mathieu,
  et~al.
\newblock Open-ended learning leads to generally capable agents.
\newblock {\em arXiv preprint arXiv:2107.12808}, 2021.

\bibitem{perez2021attention}
Jorge P{\'e}rez, Pablo Barcel{\'o}, and Javier Marinkovic.
\newblock Attention is turing-complete.
\newblock {\em Journal of Machine Learning Research}, 22(75):1--35, 2021.

\bibitem{oord2017neural}
Aaron van~den Oord, Oriol Vinyals, and Koray Kavukcuoglu.
\newblock Neural discrete representation learning.
\newblock {\em arXiv preprint arXiv:1711.00937}, 2017.

\bibitem{zhou2021informer}
Haoyi Zhou, Shanghang Zhang, Jieqi Peng, Shuai Zhang, Jianxin Li, Hui Xiong,
  and Wancai Zhang.
\newblock Informer: Beyond efficient transformer for long sequence time-series
  forecasting.
\newblock In {\em Proceedings of AAAI}, 2021.

\end{thebibliography}

\end{multicols}

\end{document}